\pdfoutput=1
\def\year{2019}\relax
\documentclass[letterpaper]{article} %DO NOT CHANGE THIS
\usepackage{aaai19}  %Required
\usepackage{times}  %Required
\usepackage{helvet}  %Required
\usepackage{courier}  %Required
\usepackage{url}  %Required
\usepackage{graphicx}  %Required
\usepackage{amsmath}
\usepackage{amssymb}
\usepackage{algorithm}
\usepackage[noend]{algpseudocode}
\usepackage{array,multirow}
\usepackage{caption, subcaption, graphicx}
\usepackage[colorinlistoftodos]{todonotes}
\usepackage{url,color,todonotes}
\graphicspath{{./pix/}}

\newcommand{\el}{\textit{et al.}}
\newcommand{\pw}{\emph{Planet Wars}}

\begin{document}
\setcounter{secnumdepth}{2}  
%\sloppy

%\author{\IEEEauthorblockN{Simon M. Lucas, Jialin Liu, Ivan Bravi, Raluca D. Gaina,
%John Woodward, Vanessa Volz and Diego Perez-Liebana}
%\IEEEauthorblockA{School of Electrical Engineering and Computer Science\\
%Queen Mary University of London\\
%London, UK\\
%Email: \{simon.lucas, jialin.liu, r.d.gaina, i.bravi, j.woodward, v.volz,diego.perez\}@qmul.ac.uk}

 \author{Simon M Lucas\textsuperscript{1},
 Jialin Liu\textsuperscript{2},
 Ivan Bravi\textsuperscript{1}, 
 Raluca D. Gaina\textsuperscript{1},\\ 
\Large{\textbf{John Woodward\textsuperscript{1},
Vanessa Volz\textsuperscript{1}, Diego Perez-Liebana\textsuperscript{1}}}\\
 \textsuperscript{1}{Queen Mary University of London, London, UK}\\
 \textsuperscript{2}{Southern University of Science and Technology, Shenzhen, China}\\
 simon.lucas@qmul.ac.uk,
liujl@sustc.edu.cn,
i.bravi@qmul.ac.uk,
r.d.gaina@qmul.ac.uk, \\
j.woodward@qmul.ac.uk,
v.volz@qmul.ac.uk
and diego.perez@qmul.ac.uk
}

\title{Efficient Evolutionary Methods for Game Agent Optimisation:\\ Model-Based is Best}

\maketitle

\begin{abstract}

This paper introduces a simple and fast variant of Planet Wars as a test-bed for statistical planning based Game AI agents, and for noisy hyper-parameter optimisation.  Planet Wars is a real-time strategy game with simple rules but complex game-play. The variant introduced in this paper is designed for speed to enable efficient experimentation, and also for a fixed action space to enable practical inter-operability with General Video Game AI agents. If we treat the game as a win-loss game (which is standard), then this leads to challenging noisy optimisation problems both in tuning agents to play the game, and in tuning game parameters. 

Here we focus on the problem of tuning an agent,
and report results using the recently developed N-Tuple Bandit Evolutionary Algorithm and a number of other
optimisers, including Sequential Model-based Algorithm Configuration (SMAC).  Results indicate that the N-Tuple Bandit Evolutionary offers competitive performance as well as insight into the effects of combinations of parameter choices.
\end{abstract}

\section{Introduction}\label{sec:intro}

%% \todo[inline]{Simon to write motivation for paper}

Games provide a rich and natural source of noisy optimisation problems.  Uncertainty arises in many ways in the design, testing and application of game-playing agents, and in the design optimisation of games to meet desired criteria.  Some games are inherently stochastic due the roll of dice, shuffle of cards or random number generation, but other sources of uncertainty include the unpredictable actions of an opponent, the use of inherently stochastic agents, and the results of human play-testing of games.

Regardless of the purpose of the optimisation, there is a natural desire to make the optimisation as efficient as possible.  This efficiency enables agents and game parameters to be optimised rapidly, enabling the use of such optimisers to be more widespread.  Extremely rapid optimisation could even spur the development of new game design tools, where a designer could be given immediate advice on the likely effects of changes to the game design, as the optimiser explores the most interesting choice of game parameters.

It is widely known in machine learning and algorithm optimisation that using a good choice of parameters can improve performance by an order of magnitude or more, compared to a poor choice.  On the game design side, well chosen parameters can be the difference between an excellent game and an unplayable one.  Within the game AI community, evolutionary algorithms are a very common choice of optimisation method \cite{yannakakis2018artificial}.  However, there are often significant advantages in using a more sophisticated model-based evolutionary approach, and
we provide an example of this below.

We designed and implemented a simplified version of Planet Wars, which is itself a relatively simple (relative to a game like StarCraft for example) but interesting real-time strategy game.  
Although Planet Wars has been used as the basis for a very successful Google AI challenge, the software behind that is complex and not
fast enough for the practical application of statistical forward planning algorithms such as Monte Carlo Tree Search and Rolling Horizon Evolution.  To enable the use of these algorithms and also allow rapid experimentation we designed and implemented a simplified version of the game.  This cut-down version provides a good benchmark for game AI agents in that it has sufficient skill-depth to expose a wide span of player abilities, but runs very quickly, at more than 10 million game ticks per second.   Evidence of skill-depth will be presented in Section~\ref{sec:res}.

The version described in this paper is as simple as we could make it
while retaining some of the essence of the game.  A fast and full-featured re-implementation of Planet Wars that adds gravity and a variety of actuators
is described by Lucas \cite{Lucas2018SpinBattle},  though that
version is an order of magnitude slower than the version described here.
In this paper we focus 
on optimising game-playing agents, and leave optimising the game 
parameters to meet specified design criteria for future work.

Another motivation behind this work to is provide an additional set of games for the General Video Game AI (GVGAI) competition framework.  While most GVGAI games to date have been implemented in the Video Game Description Language (VGDL), there are advantages to writing games in a language such as Java, as done for this paper, as noted in \cite{GVGAISurvey}.  Java provides more flexibility in designing game rules that would be difficult to express in VGDL, and also enables much faster execution of the game.  In addition to the implementation of the game, we also provide a wrapper class so that GVGAI agents see the standard GVGAI interface, and can play the game without any modification.  The Planet Wars variant 
developed here is suitable for either the single player or two player planning tracks, and has already been tested with the GVGAI framework.
With some interface wrapping it could also be used for the GVGAI learning tracks.  The development of games with strategic
depth for GVGAI and for Atari Learning Environment (ALE) agents
will provide a fresh range of challenges for general video game playing.

Having set up the game with an agent configuration problem
as a noisy optimisation problem, the main contributions
of the paper are to show the effectiveness of model-based
optimisation approaches compared to non model-based evolution, 
and also to provide insight in to how the N-Tuple Bandit Evolutionary Algorithm (NTBEA) operates.
%In this paper, we compared the NTBEA to one of the most popular hyper-parameter optimization algorithms, Sequential Model-based Algorithm Configuration (SMAC)~\cite{hutter2011sequential}, on a simpler version of the challenging Real Time Strategy game \pw~, run as a highly successful Google Game AI Challenge in 2011~\cite{fernandez2011optimizing}. The implementation of a simpler version of \pw~allows faster game simulations, while some challenging aspects of the original game are kept.

The rest of the paper is structured as follows. Section \ref{sec:noisy} presents the algorithms considered in this work for optimizing the hyper-parameters.
Section \ref{sec:planet} describes variations of the
Planet Wars game.
Section \ref{sec:prob} describes the test problem. Section \ref{sec:res} presents the experimental results and discusses the results. Section \ref{sec:concl} concludes.

\section{Hyper-Parameter Optimisation}
\label{sec:noisy}
%\todo[inline]{I think this works better as a separate section rather than dive in to this in the introduction.}

In this section we describe the main approaches to
handling hyper-parameter optimisation, specifically for noisy applications.

\subsection{Main Approaches}

The performance of an algorithm may be highly dependent on its parameter settings, which is why hyper-parameter optimisation, i.e. finding optimal parameter configurations for optimisation algorithms, is a popular topic of research. %For instance, there is a correlation between the convergence rate and the population size for population-based Evolutionary Computation (EC) techniques, e.g., Genetic Algorithms (GAs), Differential Evolution (DEs), Evolutionary Strategies (ES) and Particle Swarm Algorithms (PSAs). 
%Again, using the example of population size, Storn~\cite{storn1996usage} suggested increasing the population size for DEs with the problem dimension, while Chen~\el~\cite{chen2012large} found that large population is not always helpful in EAs.  However, when the fitness evaluation
%budget is small, having a large population size is not the solution since the fitness evaluation budget may be 
%insufficient to evaluate even
%a single generation, or place unrealistically small limits
%on the number of generations than can be evaluated.
In hyper-parameter optimisation, especially in context of real-world problems with already expensive fitness functions, the evaluation of a single parameter configuration is usually computationally expensive. For this reason, hyper-parameter optimisation algorithms often use model-based approaches \cite{hutter2011sequential,SPOT}. These algorithms train surrogate models on evaluated solutions and use the obtained information to reduce the number of evaluations required in order to avoid prohibitively long algorithm runtimes. Several methods to integrate the acquired information have been proposed \cite{Jin2011}.

The existence of noise adds another dimension to hyper-parameter optimisation, as it significantly affects the performance 
of an optimisation algorithm and may require special coping measures to be
taken, such as selective resampling as done by the N-Tuple Bandit Evolutionary Algorithm.

\subsection{N-Tuple Bandit Evolutionary Algorithm}
The N-Tuple Bandit Evolutionary Algorithm (NTBEA) was formalised by Lucas~\el~\cite{NTBEA-WCCI-2018} in 2018, though an earlier version has been used for evolving game parameter settings~\cite{kunanusont2017n}.
The NTBEA combines evolution with bandit-based sampling and
follows on from the bandit-based Random Mutation Hill Climber proposed previously~\cite{liu2017bandit}. The NTBEA was designed for hyper-parameter optimization, and has been tested on evolving new game instances by optimizing game parameter settings~\cite{kunanusont2017n}, on off-line optimization of parameter settings for game-playing AI agents on different video games~\cite{NTBEA-WCCI-2018} and on on-line learning of parameter settings for an MCTS agent on games in the General Video Game AI (GVGAI) framework~\cite{sironi2018self}.

For noisy game optimisation problems the NTBEA has a number of  attractive features:
\begin{itemize}
\item Rapid convergence to good solutions in cases of high noise and small evaluation budgets
\item Informative and intuitive model provides statistics to explain parameter choices
\item Low computation overhead compared to Gaussian Processes models, for example
\item Computation overhead scales well for large search spaces and large evaluation budgets due to efficient constant-time data structures, namely arrays or hash-maps
\item The algorithm is relatively simple, making it easy to port to and embed in new platforms
\end{itemize}

Currently there exist open source implementations of the NTBEA in Java\footnote{\url{https://github.com/SimonLucas/ntbea}} 
and in Python\footnote{\url{https://github.com/Bam4d/NTBEA}}.

%% Simon commented out these paragraphs: that's how the old
%% Bandit algorithm worked, not the NTBEA
% The whole model can be seen as a two level bandit problem. A macro-bandit has different $n$-tuples as arms. The arms ($n$-tuples) are themselves bandits with its legal value combinations as arms. An illustration example is given in Figure \ref{fig:2bandit}.

% At each iteration of the algorithm, an MAB algorithm can be applied to select the tuple to be evaluated, then another or an identical MAB algorithm can be applied to select a value combination for this tuple. When evaluating a tuple composed by $c_i$, the fitness value is not only used to update the average fitness of this tuple, but also to updates the one of the 1-tuple $(c_i)$, as well as the number of pulls.

\iffalse
\begin{figure}
\centering
\includegraphics[width=.6\linewidth]{ntuple-crop}
\caption{\label{fig:2bandit}An example of NTBEA with three $1$-tuples, two $2$-tuples and a $3$-tuple for a $3$-dimensional problem.}
\end{figure}
\fi

\subsection{Sequential Model-based Algorithm Configuration}\label{sec:smac}
For comparison, we consider one of the most popular hyper-parameter optimization algorithms, 
Sequential Model-based Algorithm Configuration (SMAC), proposed by Hutter~\el~\cite{hutter2011sequential}.

%\todo[inline]{John to write this part}

SMAC is a model-based algorithm configuration tool 
and is an extension of Sequential Model-Based Optimization (SMBO).
The algorithm alternates between constructing  predictive models and 
utilising them to determine which  parameter configurations  to choose next.
This approach is described by the authors as a balance between intensification and diversification (similar to exploitation and exploration).
%One of the advantages of this approach is that it removes the initial costly design by injecting randomly selected parameters throughout the process. 
%In addition, 
%SMAC  extends SMBO  by being able to handle 
%\emph{categorical parameters} as well as purely numerical parameters. 
%SMAC builds its response surface using random forests.
%SMAC statistically outperforms  many other parameter configuration techniques. 
%The point is made 
%by Hutter~\el~\cite{hutter2011sequential},  
%that while these gains are small when compared to SMAC, ROAR, TB-SPO [19], GGA [10], and PARAMILS,  this improvement will scale up on more costly  problem domains. 
We use the SMAC version 3 developed by the ML4AAD Group of the University of Freiburg\footnote{SMAC on GitHub: \url{https://github.com/automl/SMAC3}.} using Python and C++.

%%%%%%%%%%%%%%%%%%%%%%%%%%%%%%%%%%%%%%%%%%%%%%

\section{Planet Wars}
\label{sec:planet}

\subsection{Google AI Challenge and Dagstuhl AI Hackathon}

Planet Wars was the subject of the Google AI challenge in 2010
by the University of Waterloo in Canada ~\cite{fernandez2011optimizing}.
The game is a simple real-time strategy game that is fun for humans to play and provides
an interesting challenge for AI.  

%The are many version of the game but they adhere the following rules.

%\begin{itemize}
%\item The game is for two players (larger numbers of players would also be possible, but not considered here).
%\item The game map shows a set of planets laid out in 2D space.
%\item Each planet has a number of ships on it.
%\item Each planet is owned either by player 1, player 2, or is neutral.
%\item Each planet owned by a player spawns new ships at a rate depending on the size of the planet; large planets spawn ships more quickly than small planets.  Large planets are therefore more valuable to own.
%\item Neutral planets do not spawn new ships, but may be invaded by either player.  Deciding which neutral planets to invade is an important aspect of strategy.
%\item At each game tick, a player may move a number of ships from any planet they own.  The ships may be moved to defend a planet they already own, or to invade an enemy or neutral planet.
%\item The ships travel between the source and destination planet and take a non-zero number of game ticks to arrive.
%\item On arrival, the ships are simply added to the total if the planet is already owned by the same player, or subtracted from the total if owned by a neutral or enemy player.  If the total goes negative then the ownerhip switches to the invading player, who keeps the new total (as a positive number).
%\item The game terminates when all the planets are owned by a single player or are neutral.  In other words, a player only needs to eliminate their opponent; there is no need to occupy all the neutral planets.
%\end{itemize}

According to Buro et al.\ (section 4.1 in \cite{lucas2015artificial}), who used Planet Wars in a hackathon, the game benefits from a simple rule set and a real-time decision complexity.
%Buro et al.\ (section 4.1 in \cite{lucas2015artificial}) used Planet Wars as the basis for an AI hackathon and commented: 
%\begin{quote}
%It [Planet Wars] was chosen as the application domain for this
%Hackathon project for its simple rule set, existing programming frameworks, real-time
%decision complexity, and our curiosity about how state-of-the art AI techniques could be
%successfully applied to the game by AI experts who only have a day to implement a complete
%player.% After introducing the Planet Wars game, we describe each approach in turn, what
%worked and what did not, experimental results, and why we as a group feel that this was one
%of the best workgroup experiences we ever had.
%\end{quote}
They also observed that the 
particular implementation was rather slow and only able to play of the order of one game per second, 
which made optimising the AI players a time-consuming process.% and limited the 
%extent to which stochastic forward planning agents (Notably Monte Carlo Tree Search)
%could be applied.
%While commercial versions of Planet Wars (for example on the App Store and the Play Store) 
%often send individual ships between planets, the Google AI Challenge version 
%represented a fleet of ships as a single mothership with the number of individual ships shown
%by a number (see Figure~\ref{fig:DagPlanetWars}).

%Probably the biggest effect of showing individual ships is to make the game more visually appealing.
%However, it does also have an impact on strategy (or at least on tactical play), 
%especially if the individual ships implement collision detection to prevent overlapping in space;
%the effect is to reduce the rate at which planets can be invaded, since the ships arrive
%one-by-one rather than forcing a step change on the planet ownerships.

%\begin{figure}
%\centering
%\includegraphics[width=.8\linewidth]{PlanetWarsDagstuhl}
%\caption{\label{fig:DagPlanetWars}A screenshot of the version of Planet Wars used for a recent Dagstuhl AI Mini-Hackathon.  The planets belong to each player are filled with the player's colour (blue or green); neutral planets are shown in grey, and do not grow ships.  Each triangle shows a fleet in transit between a source and destination, with the number of ships in that fleet shown as a number (not legible in this figure).}
%\end{figure}

\subsection{Fast Planet Wars Variants}

The approach we took to developing variants of Planet Wars was to strip it back to the bare
minimum of essential features which would still allow the game to be recognisable
as a planet invasion game. We began with the following aims:

\begin{itemize}
\item The game should be fast, allowing rapid copying of the game state, and even more rapid advancing of the game state given a set of selected actions (i.e. the nextState function).  We aim to run the game at more than one-million ticks per second on a typical model laptop.  This makes it suitable for testing statistical forward planning algorithms.
\item Make the game easily scalable to a large number of planets, without changing the structure of the agents or making the game too slow.  Ideally the computational cost of the next-state function should be independent of the number of planets, or at worst be linear in the number of planets.
\item Ensure the game has significant skill-depth (this can be estimated by the span of Elo ratings (or more simply, league table scores), between strong and weak agents). 
\item Make the code simple and easily extensible to encourage others to copy and adapt.
\item Retain the simultaneous real-time aspect of the game.
\item Make the game playable with a small fixed-size action space, suitable for General Video Game AI agents, or Atari Learning Environment (ALE) agents.  
\end{itemize}

%Regarding the fixed-size action space, the original Planet Wars games involve creating lists of orders at each term, with each order specifying the source and destination planets plus number of ships to transfer.  Current general video game playing agents are not designed to produce lists of orders, though this in itself represents an interesting future challenge.

\begin{figure}
\centering
\includegraphics[width=.8\linewidth]{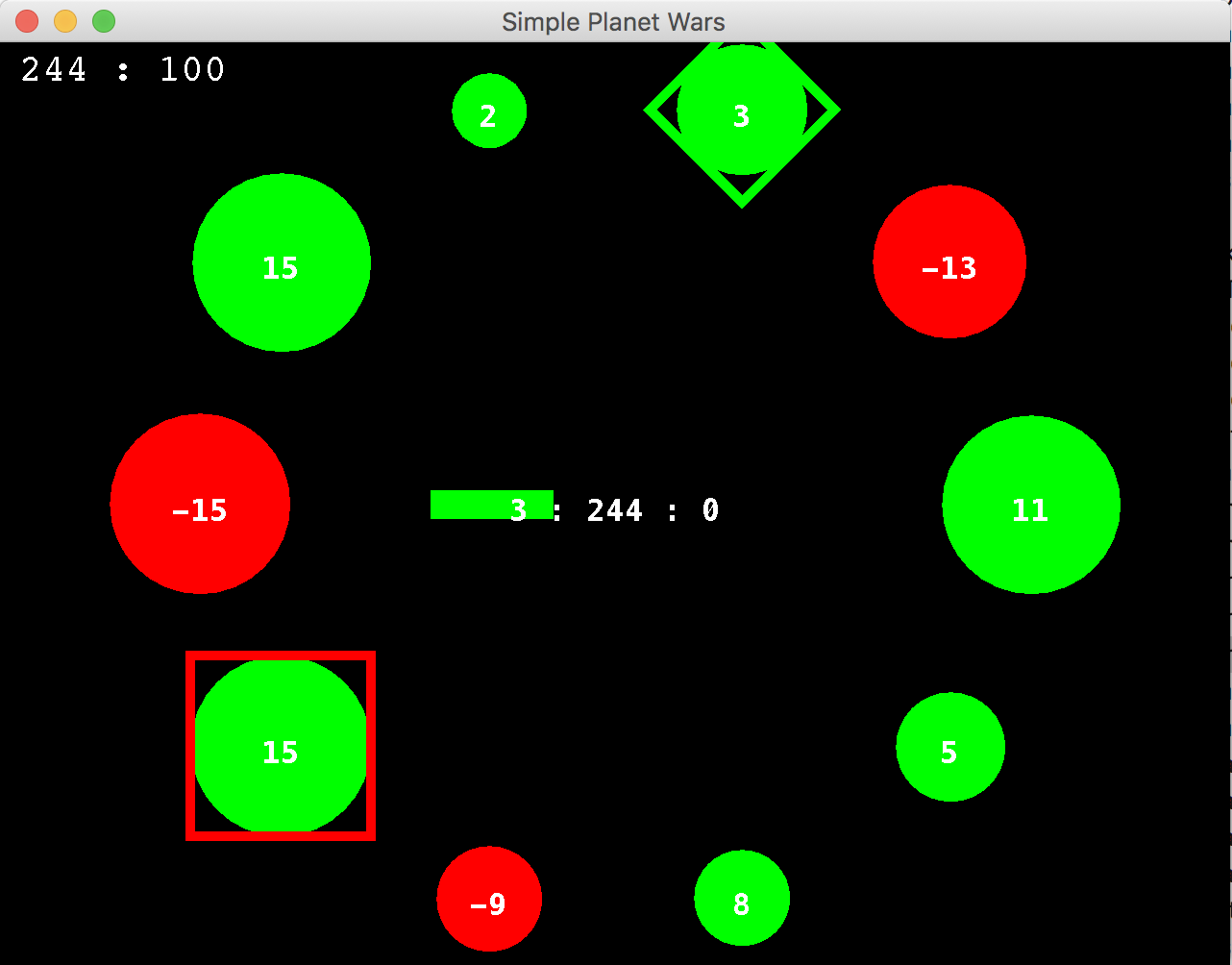}
\caption{\label{fig:SimplePlanetWars} A simplified version of planet wars made for speed. 
This is a competitive two player game played by the green player versus the red player.  In this
version there are no neutral planets.  All transfers between planets go via a player's respective
buffer in the middle of the area, and are transferred between the buffer and the planet of focus, identified by the green diamond and the red square.   The aim for each player is to acquire the most ships within a specified
time limit or to win the game by occupying all the planets.}
\end{figure}

A screen shot of our simplified Planet Wars Game is shown in Figure~\ref{fig:SimplePlanetWars},
and was first described \pw~by Lucas~\el~\cite{NTBEA-WCCI-2018}. 
The game well exceeds our target
speed and runs at more than $10$ million game ticks per second. 

We first reiterate the main features of the 
game from \cite{NTBEA-WCCI-2018} before describing some
aspects in more detail:

\begin{itemize}
\item There are no neutral planets: the ships on each planet are either owned by player 1 or player 2. 
\item At each game tick, a player moves by shifting ships to or from a player's ship buffer, or by moving its current \emph{planet of focus}.
\item When a player transfers ships it is always between its buffer and the current planet of focus.
\item At each game tick the score for each player is the sum of all the ships on each planet it owns, plus the ships stored in its buffer.  We have two versions of the game: the easiest for the planning agents, and the one used for most of the experiments in this paper, also adds in the ships in a player's buffer to its score, a more deceptive version of the game does not include this.  
\end{itemize}

% In future work we plan to build up from these minimal features
% and to evaluate the difference that each element makes in terms of
% how it affects the skill depth of a potentially large set of agents.  

Our intuition is that two of the elements missing from the
cut-down version will most likely significantly reduce the skill-depth: these are the time-delay between the ships leaving their origin and arriving 
at their destination, and the neutral planets.  Neutral
planets pose an interesting dilemma for a player: invading them
wastes a player's ships, unlike invading the opponents planets
where the ships lost directly deplete the opponent's total.  But invading neutral planets is also necessary in order to grow a
player's ship producing capacity.  Therefore a common ploy is
to wait for the opponent to invade a neutral planet, then issue
an immediate strike on it.  Hence, it will be interesting future work to
compare the skill-depth of the version described in this paper with the
version described in \cite{Lucas2018SpinBattle}.

\subsection{Game Play}

At each game step a player executes one of 5 actions.  Do nothing, move planet of focus (clockwise or anti-clockwise), or attempt to move planets from buffer to planet of focus, or vice versa.

To make the agent evaluation process as efficient as possible
the game used in all our experiments was played for a fixed budget
of $200$ game ticks, as most games were already decided by this point.
%In general planet wars games tend to go
%through an interesting phase where pivotal battles are fought
%and the outcome is highly uncertain, and then transition
%into a less interesting phase when one of the players dominates
%and the outcome is almost certain.  Many classic games such as chess
%avoid this by having highly non-linear game play, where a player
%who is behind on pieces may still win due to better positional play.
%A future aim of this work is to develop and / or evolve new variants of this cut-down game that retain the possibility for the underdog to come
%from behind.  Ideas that may aid this are the concept of a planet being liable to explode when it reaches a certain limit of ships, and also allowing mid-flight collisions where a small fleet name destroy
%a large fleet when in transit.
The objective of each player is to have the higher score when the game
ends%finishes (after the set number of ticks, in this case 200),
, which is calculated as the sum of all the ships on each of their planets plus the ships stored in their buffer.

In addition to the agent interface, the game is also playable via the arrow-keys on a keyboard.
The lead author has played many games against the best evolved agents
in this study using the best configurations specified below,
and is rarely able to beat them when playing at one second per
move.  However, the circular layout of the planets is confusing
for left/right key control (just as playing Atari's Tempest using a keyboard is confusing).  
% A long term aim of the work is to produce 
% variants which are enjoyable for human players.

% The outcome of this game is not symmetric, the sum of both players' scores may be different from 0. This is a general-sum game.

% \pw~is a simple but challenging Real Time Strategy game
% that was run as a highly successful Google Game AI Challenge in 2011 by the University of Waterloo in Canada~\cite{fernandez2011optimizing}.

%%%%%%%%%%%%%%%%%%%%%%%%%%%%%%%%%%%%%%%%%%%%%%
\section{Optimising Rolling Horizon Evolutionary Planning Agents}\label{sec:prob}

We compared the NTBEA and SMAC (presented in Section \ref{sec:noisy}), as well as several non-model based evolutionary algorithms, on optimizing hyper-parameters of a game playing agent on the Fast Planet Wars game.
%described above, and also compared a number of other evolutionary algorithms.
% stochastic game \pw. 
% Section \ref{sec:game} describe the game. Section \ref{sec:agent} describe the game playing agent and its hyperparameters.

\subsection{Agents}\label{sec:agent}
In this work, we optimise the parameters of a rolling horizon $(1+1)$-Evolutionary Algorithm, denoted as RHEA in the rest of the paper.
In the RHEA agent, the individual is an action sequence of a fixed horizon.  The algorithm is initialised by creating a population of random actions sequences or rollouts by sampling uniform randomly from
the available set of actions.  The choice of evolutionary algorithm optimiser is one of the parameters of the RHEA agent.  For all these experiments we used a random mutation hill climber, also known as
a $(1+1)$ EA.  In this case the algorithm is initialised with a single random rollout.

%. In the optimisation procedure, the individual shifts its action sequences by one position and fills its last position by a randomly selected action, instead of randomly initialising a whole new sequence.

The key parameters to be optimised for this agent and their legal values are summarised in Table \ref{tab:space}.
\begin{table} [!t]
\centering
\caption{\label{tab:space}Search space of the parameter settings.}
\begin{tabular}{ccc}
\hline
Variable & Type & Legal values \\
\hline
$nbMutatedPoints$ & Integer & $0.0, 1.0, 2.0, 3.0$ \\
$flipAtLeastOneBit$ & boolean & $false, true$ \\
$useShiftBuffer$ & boolean & $false, true$ \\
$nbResamples$ & Integer & $1, 2, 3$\\
$sequenceLength$ & Integer & $5, 10, 15, 20, 25, 30$ \\
\hline
\end{tabular}
\end{table} 
The notations are detailed as follows:
\begin{itemize}
\item $nbMutatedPoints$ defines the mutation probability, how likely a mutation occurs at every dimension, by dividing it by the number of dimensions (in this case the number of dimensions is the same as the sequence length)
\item $flipAtLeastOneBit$ indicates if at least one mutation should occur at each time;
\item $useShiftBuffer$ enabled or not: if disabled, at any time step $t+1$, the initial individual is reset to random, otherwise, the individual at time step $t$ shifts its action sequence forward and fills the last position by a random action;
\item $nbResamples$ defines how many time the individual (i.e.\ the action sequence) is re-evaluated as the game is stochastic;
\item $sequenceLength$ defines the planning horizon (i.e.\ the length of each action sequence / rollout).
\end{itemize}
% 3     1     1     0     2

As \pw~is a two-player game without win or loss, only the game scores for both players are reported at the end of a game. We define the following evaluation function for the agent, assuming optimizing for player 1:
\begin{equation}
    f(player_1,player_2)= 
\begin{cases}
    1,& \text{if } score_1 > score_2\\
    -1,              & \text{otherwise}
\end{cases}
\label{eq:fitness}
\end{equation}
where $score_1$ and $score_2$ denotes the final scores obtained by $player_1$ and $player_2$.
The RHEA agent aims at maximising the game score from the perspective of
its player.  Since the player with the largest number of ships wins, 
this is a sensible objective to optimise.

% \todo[inline]{check with simon}

\begin{figure}
\centering
\includegraphics[width=.8\linewidth]{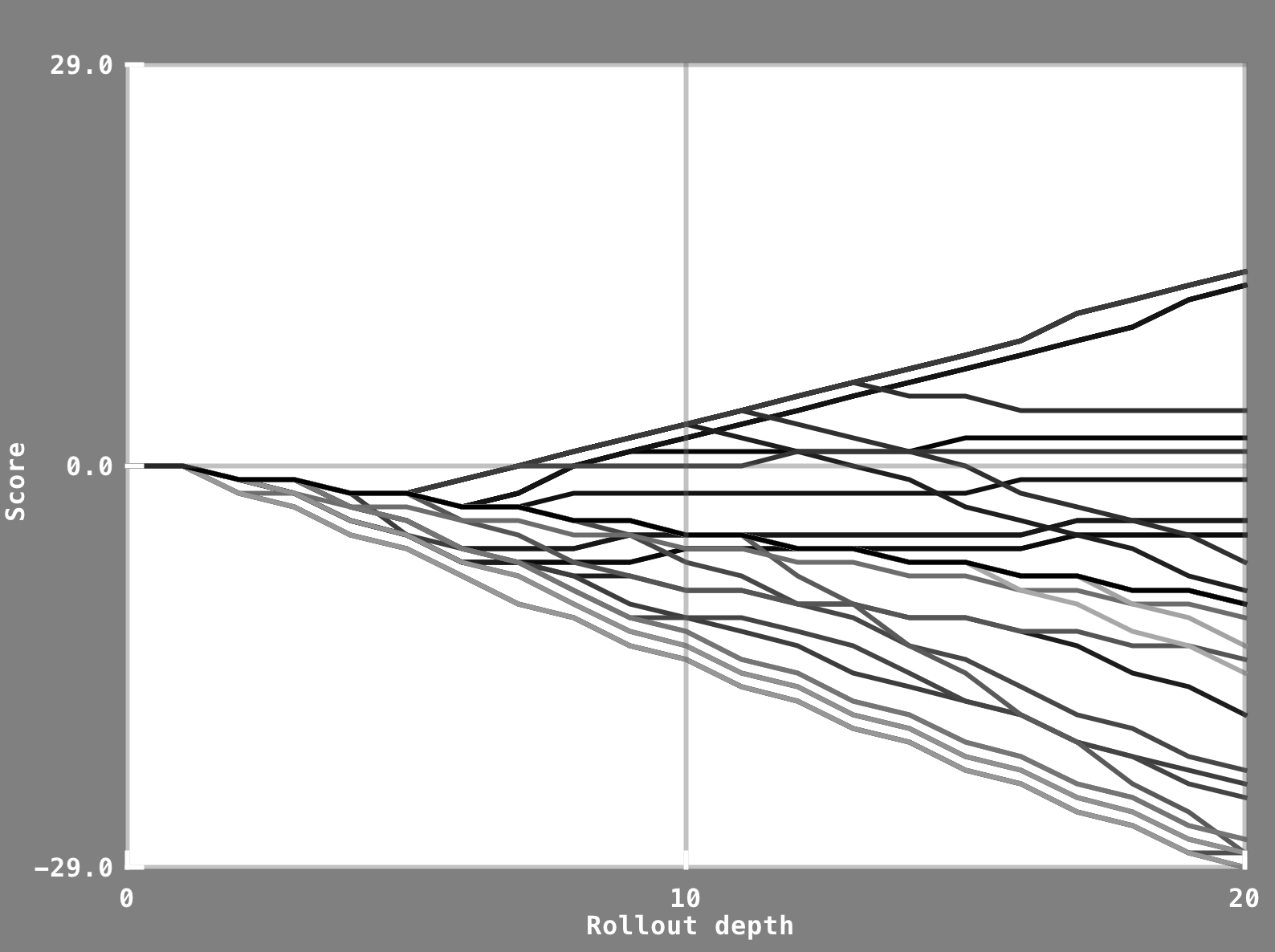}
\caption{\label{fig:IncBufferInScore} The graph provides insight into a rolling horizon evolutionary algorithm playing Planet Wars.  The RHEA agent has a rollout length  (horizon) of 20, and this variant of the game includes the ships on their buffer in each agent's score. }
\end{figure}

\begin{figure}
\centering
\includegraphics[width=.8\linewidth]{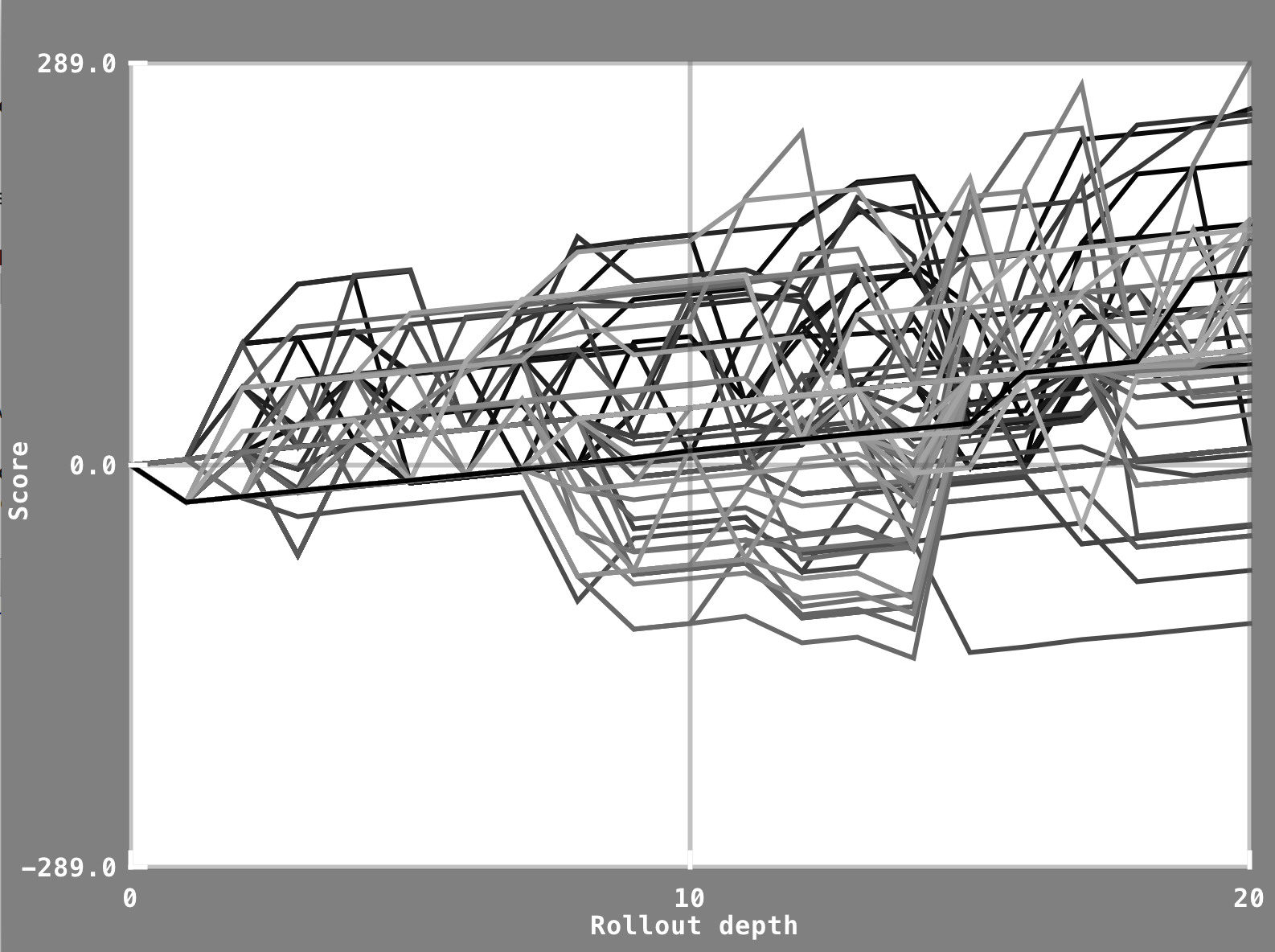}
\caption{\label{fig:NoBufferInScore} This graph is similar to the one in figure~\ref{fig:IncBufferInScore} except that the game score does not include the ships in each buffer, meaning that agents need to see beyond the apparent loss in score that moving ships to their buffer would entail.  This causes short-term agents to perform very badly compared to the previous version of the game. }
\end{figure}

\begin{figure}
\centering
\includegraphics[width=.48\linewidth]{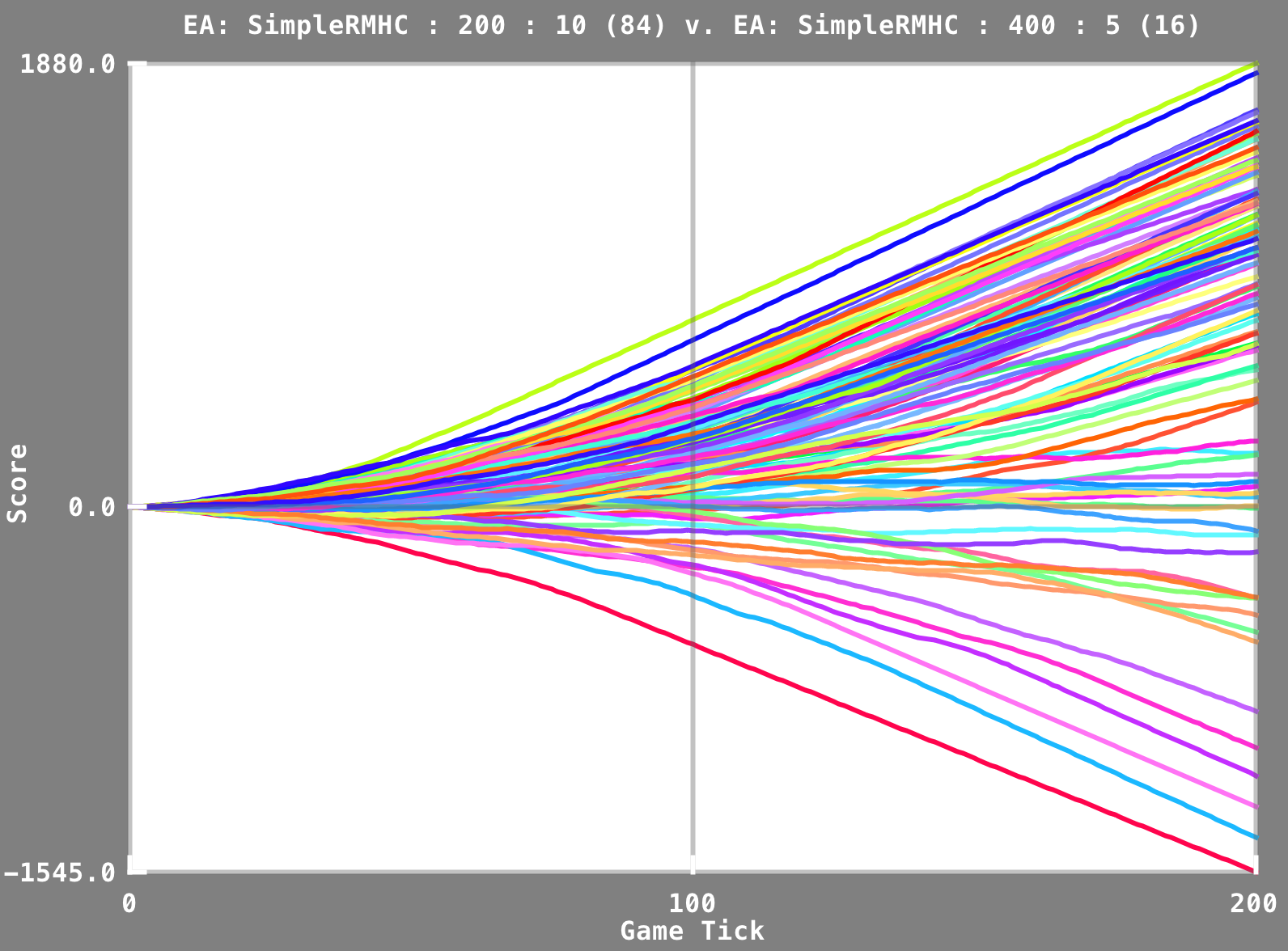}
\includegraphics[width=.48\linewidth]{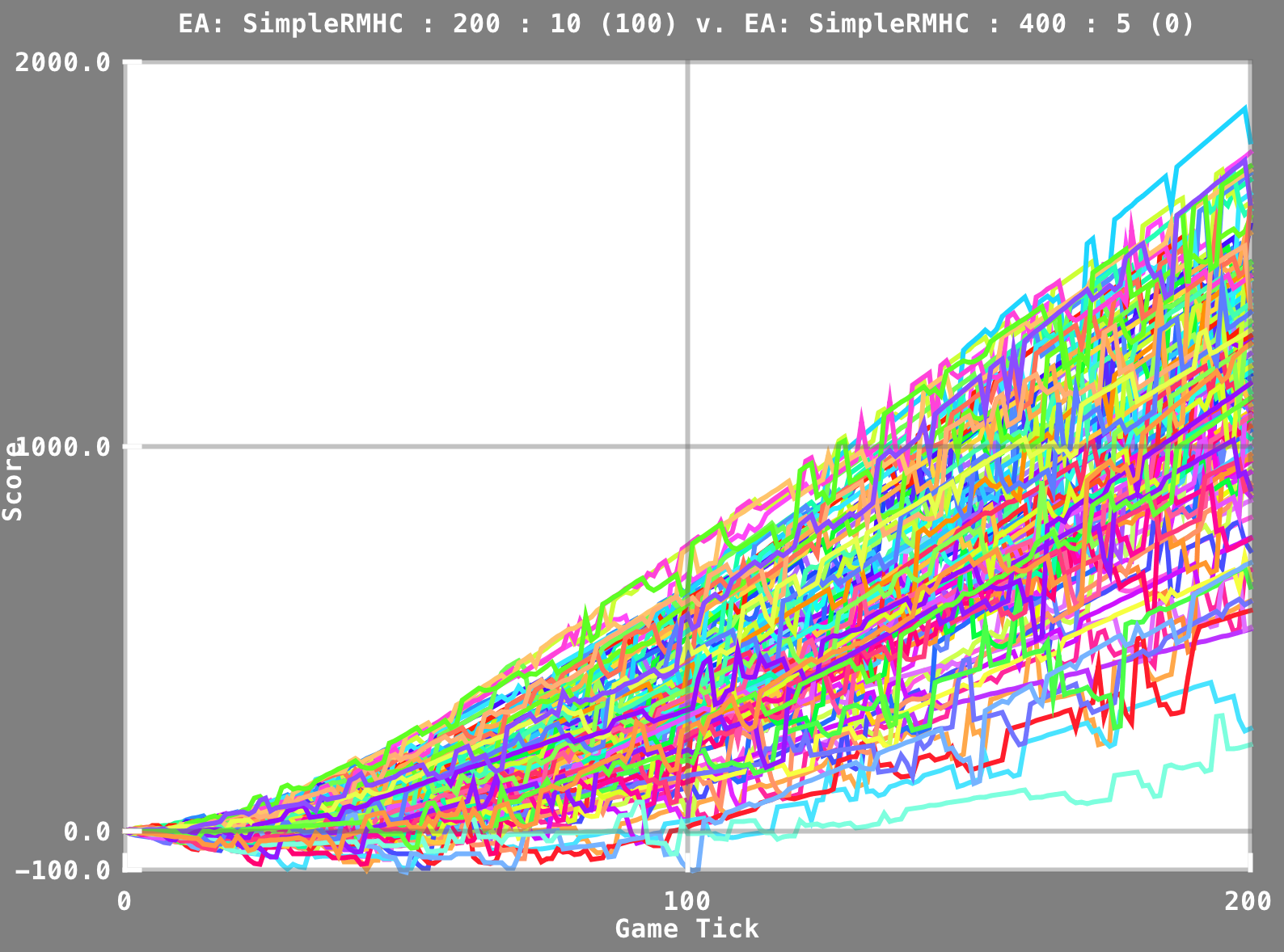}

\caption{\label{fig:GameScoreIncBuffer} The figure shows a medium-term planning agent (200 rollouts of length 10) playing 100 games against a short term agent (400 rollouts of length 5).  The number of rollouts per game state is adjusted to keep the tick-budget per action constant at $2,000$. Each line shows how the score developed over each game tick. Left: Buffers are included in the score, Right: Buffers not included. With the buffer, we observe the short-term agent being convincingly beaten, but still winning $16$ games. Without the buffer, the short-term agent lost all 100 games. This illustrates how small changes to the game can
have significant effects on the agents that play it.}

%\centering
%\includegraphics[width=.8\linewidth]{GameScoreNoBuffer}
%\caption{\label{fig:GameScoreNoBuffer} This is the same as figure~\ref{fig:GameScoreIncBuffer} except that the buffers are not included in the game score.
%This makes it especially hard for the short-term agent, which on this occasion lost all 100 games.
%This illustrates how small changes to the game can
%have significant effects on the agents that play it.}
\end{figure}

To give an idea of what each agent sees during its rollouts, we plot the score
difference from the current game state to 20 game ticks ahead for a typical game-state.
Figure~\ref{fig:IncBufferInScore} shows the results of the rollouts when
we include the ships in the buffers in each player's score.  To give an
idea of how simple changes to the game can affect an agent's view of
the world, Figure~\ref{fig:NoBufferInScore} shows how the score typically
varies when the buffers are not included in the score.  Note that this
simple change makes the game noisier and more deceptive,
and harder for the short-term agent to play.
Figure~\ref{fig:GameScoreIncBuffer} shows the game score
traces when playing these two agents against each other
for these two game variants.  Not including the buffered ships
in the score has a significant effect both on the game outcomes 
(the short-term agent now loses all 100 games) and
the smoothness of the score trajectories.

%\subsection{Micro-analysis of Game Variants on Agent Decision Making}

%The idea here is to look at changing rules during the playout of a game (Euro Games already do this for real) and observe the effects on the decisions made by the agents.

%This will give interesting additional insight in to the behaviour of agents at particular points in a game.  

%To be clear: we can take a game state, and apply different agents to make a decision on what to do next.

%Then: we change the rules, and re-evalaute.  The question is: do the agents choose the same actions as before?  This gives insight in to the significance of the rule change with respect to a set of agents.

%In the example of whether the score is included in to the shift buffer or not, not including the score would have a devastating effect on a 1-ply look-ahead agent, which would never see beyond the immediate loss to appreciate the long-term investment value of moving ships to its buffer.

%%%%%%%%%%%%%%%%%%%%%%%%%%%%%%%%%%%%%%%%%%%%%%
\section{Results and discussion}\label{sec:res}

In this section we describe the experimental settings followed by 
a more detailed analysis of the optimisation problem. The comparison with
other evolutionary algorithms is presented in Section~\ref{sec:comparison}.
%and clearly shows NTBEA and SMAC to offer the best performance.

\subsection{Experimental settings}\label{sec:setting}
The game is played by a RHEA with settings selected by an optimizer%, NTBEA or SMAC, denoted as $A_{NTBEA}$ and $A_{SMAC}$, 
 versus a RHEA agent with manual tuned parameter setting $(1,true,true,1,5)$ without knowledge about the fitness landscape% as shown in Figure \ref{fig:sort}, 
 denoted as $A_{fixed}$. At every game tick, each of the agents has $2,000$ forward model calls as its simulation budget. The fitness value reported for a solution is $f(A_{tuned},A_{fixed})$ with $f$ defined in \eqref{eq:fitness}.

%According to \eqref{eq:fitness}, the fitness of the hyper-optimizer is defined as:
%\begin{equation}
%    f(A_{tuned},A_{fixed})= 
%\begin{cases}
%    1,& \text{if } score_{tuned} > score_{A_{fixed}}\\
%    -1,              & \text{otherwise}
%\end{cases}
%\end{equation}
%where $A_{tuned}$ is either $A_{NTBEA}$ or $A_{SMAC}$ and $score_{tuned}$ refers to the score obtained by this agent; and $score_{A_{fixed}}$ denotes the score obtained by the fixed agent.

The opponent model used by each agent was fixed for all the hyper-parameter optimisation experiments reported below.  
%There is one more parameter of the RHEA agent: the opponent model used.
%This was fixed for all the hyper-parameter optimisation experiments reported below.  
%When a RHEA agent plays a two-player game, it inevitably makes assumptions about how its opponent will
%act.  These assumptions are encoded in the opponent model.
In order to identify the baseline performance, we used an agent that does not act at all
(i.e. returns a \emph{do nothing} action each time)
%This was fixed to be a Do Nothing agent (for every action it selects NO\_OP)
both for the agent being tuned, and for its fixed opponent. 
For the analysis visualised in Figure~\ref{fig:GameScoreIncBuffer} we used a random agent instead.

We used the default settings for each of the optimisation algorithms tested. In addition, we experimented with the exploration settings (parameters $k$ and $\epsilon$) in NTBEA in order to gain a better understanding of the algorithm.  The results are reported in Section~\ref{sec:comparison}.
%, where each agent used a uniform random agent for its opponent model (this was found to provide a more dramatic distinction between the medium and short-term planning agents) depending on whether the
%buffers were included in the scores or not.

\subsection{Fitness Distribution}

\begin{figure}
\centering
\includegraphics[width=1\linewidth]{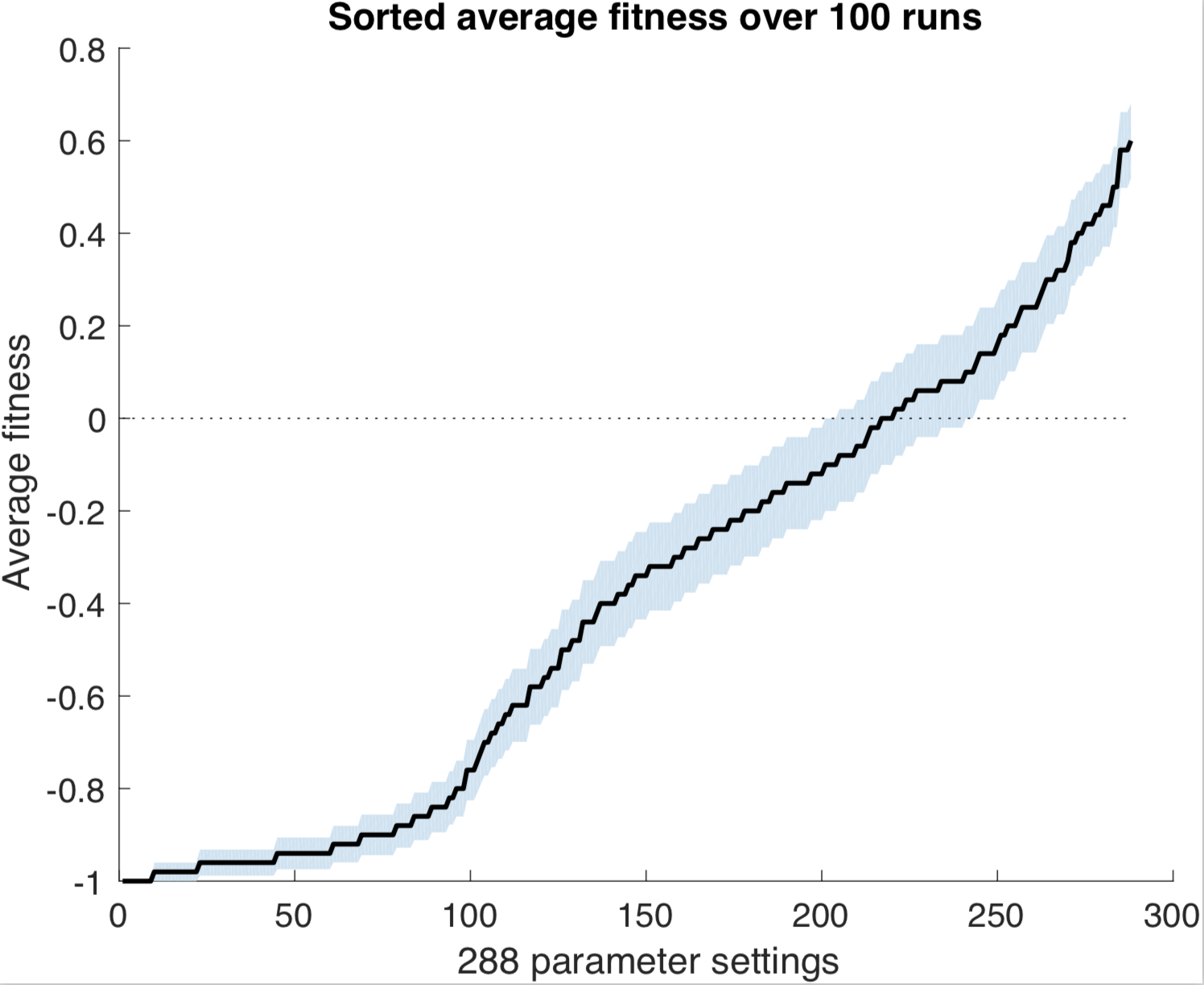}
\caption{\label{fig:sort}Sorted average fitness obtained by the agent using the 288 legal parameter settings. Each of the parameter settings is tested 100 times. Higher score fitness refers to being stronger than the opponent.
The dashed horizontal line refers to when the parameterised player has identical performance to the fixed opponent. }
\end{figure}

Figure \ref{fig:sort} illustrates the performance of the agent using the $288$ possible parameter settings described in Table~\ref{tab:space}, sorted by the average fitness value over $100$ trials (i.e.\ the RHEA agent with each set of parameters player 100 games against a fixed opponent to evaluate its fitness).
The blue shadow shows one standard error either side of the mean. Due to the inherent noise in the evaluation 
process, the parameters that appear as optimal could vary a
bit from run to run, but there are very clear differences between
different sets of players, with the best parameter settings performing much better than the worst ones.  
We consider the fact that different agents play the game with a range of clearly separable abilities as evidence of skill-depth.

%The error bars could be tightened by making more evaluations,
%but the results would not be radically
%different (we can say this having run tens of experiments in the process of writing this paper and having consistently seen a small number of parameter combinations dominate the other ones).

The best parameter setting, according to the 100 tests made of every point in the search space is $(3, true, true, 1, 15)$ with average fitness $0.6$.
As shown in Figure \ref{fig:sort}, there is a clear difference in performance between many of the different parameter settings,
though also some which are not clearly separated from each other.

\subsection{Observing the N-Tuple Statistics}

%To study the impact of parameters on the agent's performance, we repeat $100$ games played by a RHEA with a specific setting and the same RHEA agent but with manually tuned parameter setting, denoted as $A_{fixed}$. At every game tick, each of the agents has $2,000$ forward model calls as its simulation budget.

% To gain insight in to the effect of different parameter settings
% on the RHEA agent, we evaluated all 288 combinations of parameters 100 times each.

\begin{figure}
\centering
\includegraphics[width=1\linewidth]{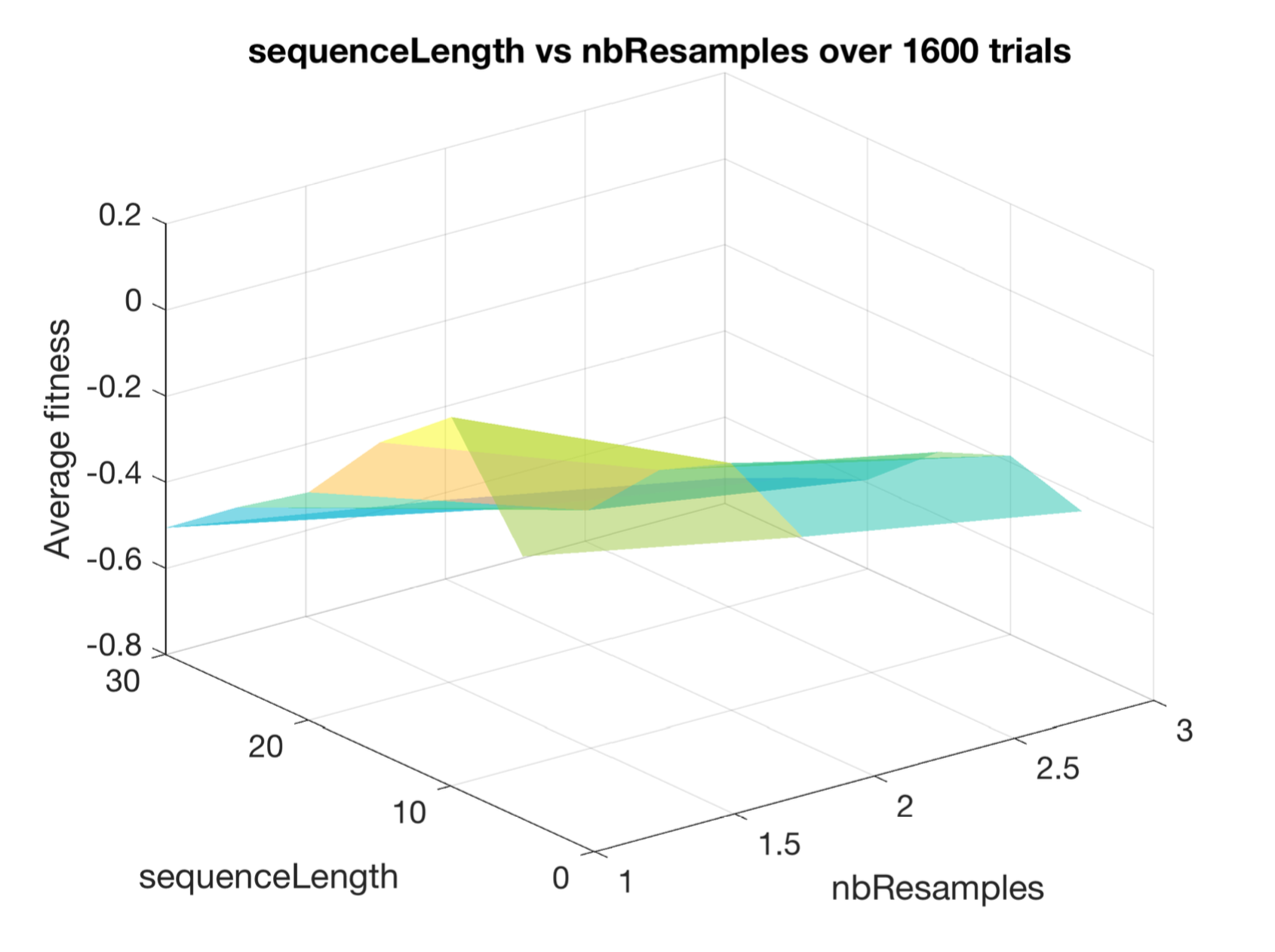}\\
\includegraphics[width=1\linewidth]{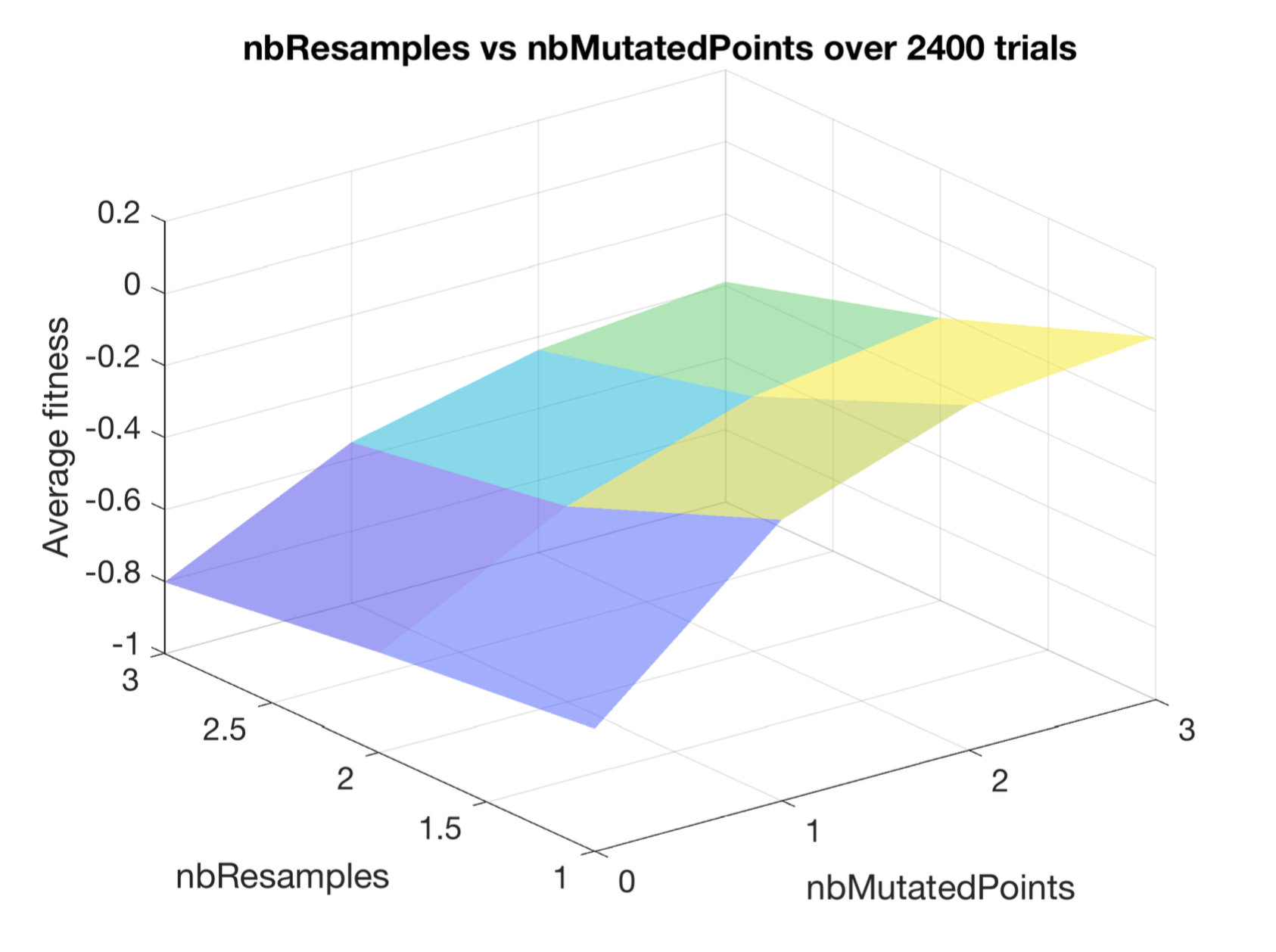}\\
\includegraphics[width=1\linewidth]{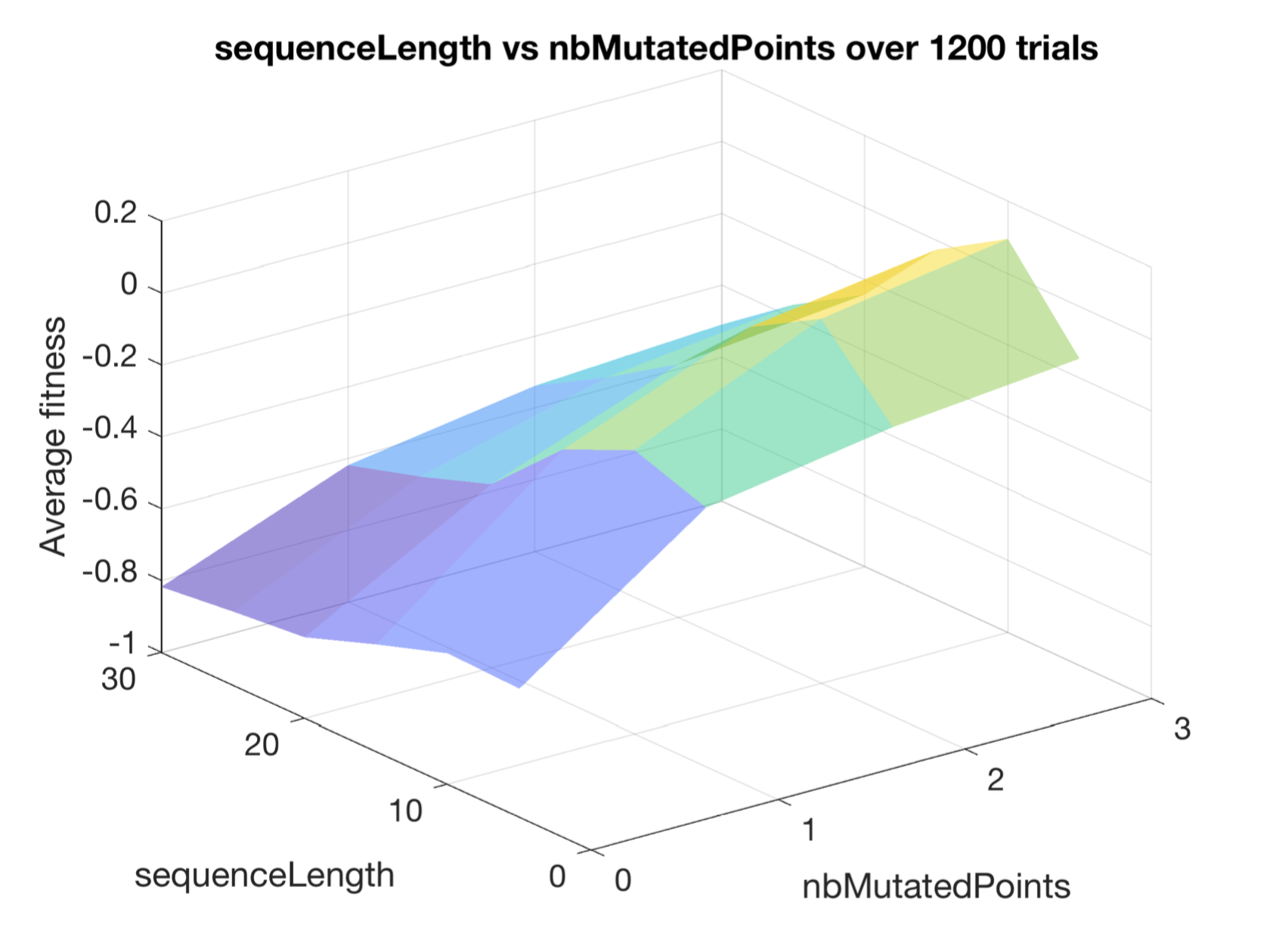}
\caption{\label{fig:2tuples}The impact of number of resamplings, mutation rate and the planning horizon (sequence length). z-axis shows the average fitness when fixing values for two of the parameters, as shown in x-axis and y-axis.}
\end{figure}

Figure~\ref{fig:2tuples} shows how fitness varies 
across a number of 2-tuple parameter choices, again
when averaged over all 288 points in the search space,
each evaluated 100 times.  Clearly there are dependencies between
the parameter combinations.  Note how most pairwise combinations
have an average value of below zero (averaged over all points sampled).
This is due to the fact that most parameter settings (more than 200 of the 288 possible as shown in Figure~\ref{fig:sort}) score below
zero.
% and the idea is that the NTBEA
%is able to exploit these during its search (as will SMAC
%using its random forest based sub-space sampling).

\begin{figure}[h]
\centering
\includegraphics[width=1\linewidth]{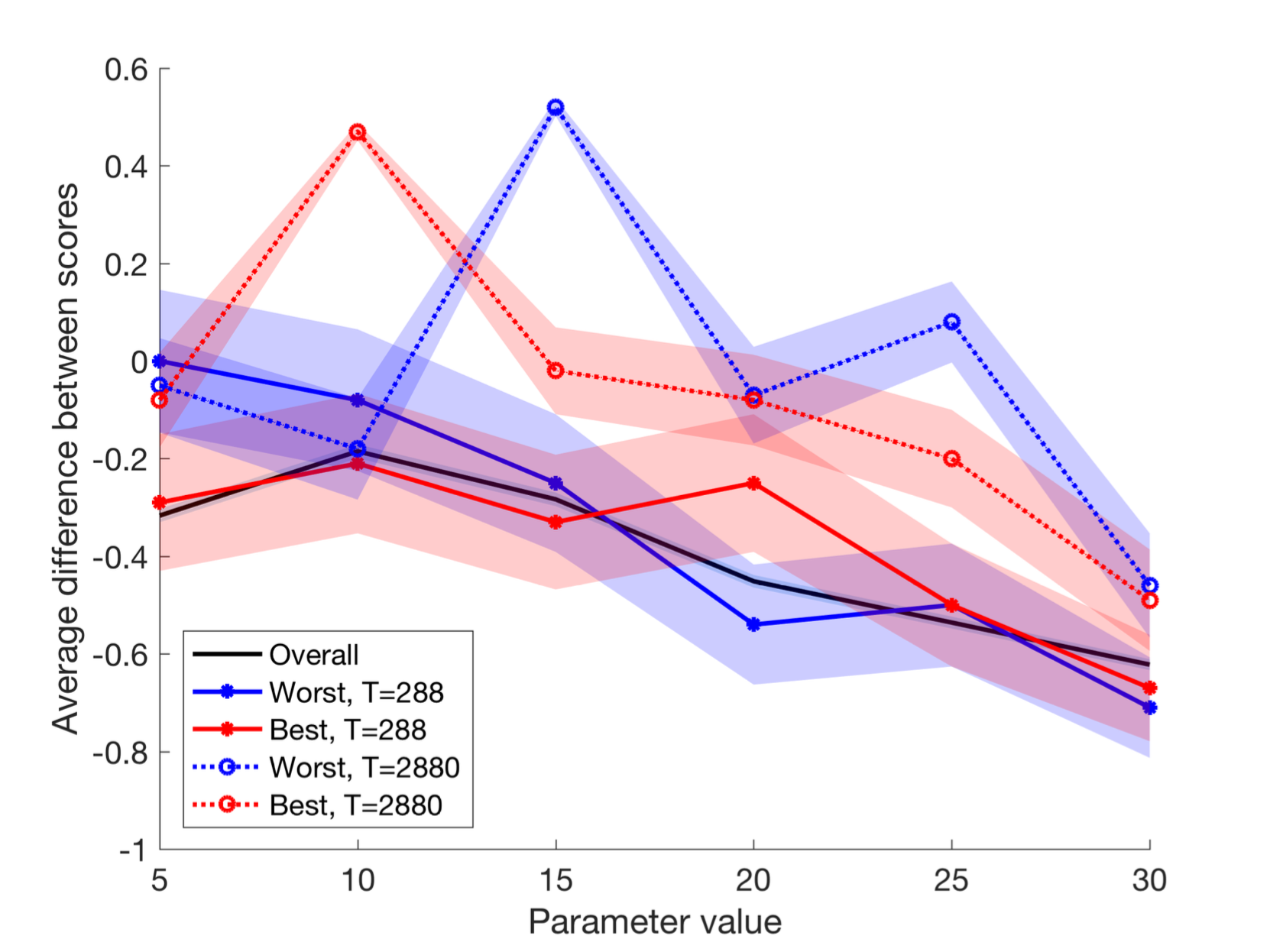}
\caption{\label{fig:ntbeaseq}Average fitness for different sequence length evaluated by the NTBEA optimizer. The black solid curve shows the average fitness over 100 repetitions of games of the agent using all the 288 possible parameter settings, and is used as a reference. The solid and dashed red curves show the results of one of the successful trials (optimum found) using $T=288$ and $T=2880$ as budget. The solid and dashed blue curves show the results of the worst trial using $T=288$ and $T=2880$ as budget.}
\end{figure}

To understand the estimation of parameters by NTBEA optimizer, we plot in Figure \ref{fig:ntbeaseq} the average fitness for different sequence lengths evaluated by the NTBEA in the worst optimisation trial and a randomly selected successful trail (optimum found) using a low budget (288 game evaluations) and a higher budget (2880 game evaluations).
As reference, the average fitness values for different sequence length over 100 repetitions of games of the agent using all the 288 possible parameter settings are illustrated using the black solid curve. 

Interestingly, the graph shows big differences in how fitness varies
with respect to sequence length depending on how the space has been sampled,
with some runs even incorrectly indicating that a length of 5 is best.
The best solutions involve a sequence length of either 10 or 15,
and this usually emerges from the statistics given enough samples.
Note though that the large sample budget runs with the high peaks
differ greatly from the overall statistics (black line).

The individual runs show how the NTBEA
samples fitter regions of the space over the course of 
a run, which is especially clear in the large positive
spikes for the $2880$ budget. 

%It is also clear that on different
%runs it may converge to different solutions.  

%More work
%is needed to understand how to tune the parameters of the NTBEA,
%but we made some effort at this for table~\ref{tab:compare},
%and were able to observe and correct a particular pathology
%of where the algorithm attempts to sample every point in
%the space before applying more selectivity.

With sufficient budget the NTBEA is able to provide a better estimation of the performance achieved by different parameter values,
and note how tight the error bars have become
for the best sequence lengths of 10 or 15 for the large 2880
budget runs, compared the error bars for worse performing sequence
lengths on those runs.

\subsection{Performance Comparison}
\label{sec:comparison}

\begin{table}[h]
\centering
\caption{\label{tab:compare}Comparison of different optimisation algorithms given $288$ game evaluations as budget, except where stated (5x and 10x indicate 5 times and 10 times the standard budget respectively; this was only used for CMA-ES.)}
\begin{tabular}{cc}
Algorithm & Avg. $\pm$ StdErr \\%& Find Opt.\\
\hline
RMHC(1) & -0.29 $\pm$ 0.01 \\ %& 0 \\
RMHC(5) & 0.01 $\pm$ 0.01 \\ %& 0 \\
CMA-ES(1x) & -0.12 $\pm$ 0.04 \\
UH-CMA-ES(1x) & -0.09 $\pm$ 0.04 \\
SGA & 0.03 $\pm$ 0.01 \\ %& 1 \\
CMA-ES(5x) & 0.29 $\pm$ 0.02 \\
UH-CMA-ES(5x) & 0.32 $\pm$ 0.03 \\
SWcGA & 0.36 $\pm$ 0.01 \\ % & 0 \\
SMEDA & 0.38 $\pm$ 0.01 \\ % &  \bf{16} \\ 
NTBEA(1,2,5)- & 0.41 $\pm$ 0.01 \\%& 13 \\
UH-CMA-ES(10x) & 0.44 $\pm$ 0.02 \\
CMA-ES(10x) & \bf0.48 $\pm$ 0.02 \\
SMAC & \bf{0.49} $\pm$ 0.01\\
NTBEA(1) & \bf{0.50 $\pm$ 0.01} \\%& 13 \\
NTBEA(1,2) & \bf{0.51 $\pm$ 0.01} \\%& 13 \\
NTBEA(1,2,5)$+$ & \bf{0.51} $\pm$ 0.01 \\%& 13 \\
\end{tabular}
\end{table}

The performance results of several evolutionary algorithms are shown in Table~\ref{tab:compare}, with brief notes below.
%We also ran the algorithms that we have implementations for 
%on further tests, with the results shown in Table~\ref{tab:compare}.

\begin{itemize}
\item 
The Random Mutation Hill Climber (RMHC) is the simplest 
evolutionary algorithm, also called a (1+1) EA.  Despite its simplicity it often performs well.  In this case though it performs poorly due to the high levels of noise.
RMHC(1) uses no resampling, while RMHC(5) resamples each candidate 5
times to alleviate the noise, but at the cost of wasted evaluations
(the total sample budget was limited to 288).
\item 
The Simple Genetic Algorithm (SGA) also performs poorly, in this case
due to its inefficient use of the small sample budget.

\item We included CMA-ES due to its high performance across a wide range
of problems.  CMA-ES operates in continuous search spaces, so
to apply it to these discrete problems we used a box constraint 
in the range 0 to 1 in each dimension, and discretized the continuous value
in each dimension by dividing into equal size ranges within the unit interval.
We also tried the CMA-ES with uncertainty handling \cite{Hansen-CMA-ES-UH}
which is meant to be better suited to noisy problems, but did not improve on
standard CMA-ES, perhaps due to the low sampling budget.
CMA-ES(5x) and CMA-ES(10x) indicate that we modified the fitness
function to use 5 times or 10 times resampling: these used up to 10 times
larger sampling budget (10x indicates that 2,880 games were played during
the optimisation). 

\item Sliding Window Compact GA (SWcGA) and SMEDA (Sliding Mean EDA) are new versions of
the Compact GA (cGA) designed to be more sample efficient by incorporating a sliding window of candidate solutions.  The SWcGA is described in ~\cite{lucas2017efficient}. They
both use models to improve sample efficiency, though the models 
only estimate which parameter values are most likely to be optimal.
In contrast, NTBEA and SMAC estimate the expected fitness of 
each parameter setting, and NTBEA goes one step further to also
estimate the variances of each parameter setting.
\end{itemize}

All model based
approaches greatly outperformed the non-model based alternatives
in this study. 
All the approaches with results on boldface are better than
the non-bold ones with statistical significance, but are not significantly
different to each other.  We ran several version of the NTBEA with different
tuning.  The numbers in parentheses show the n-tuples which were used
for each configuration of the NTBEA.  The poor performance of the NTBEA(1,2,5)$+$
is due to a particular problem we observed.  Depending on the particular 
combination of $k$ and $\epsilon$ it is possible for the NTBEA to
attempt to sample every point in the search space at least once.
This is not a problem for the smaller n-tuples, and for NTBEA(1,2,5)$+$
we fixed it by setting $\epsilon$ to 0.5 and $k$ to 1.0.  See \cite{NTBEA-WCCI-2018} for
an explanation of these parameters.

%%%%%%%%%%%%%%%%%%%%%%%%%%%%%%%%%%%%%%%%%%%%%%
\section{Conclusion and further work}\label{sec:concl}

In this paper we explored noisy optimisation in the context of a simple and fast version of Planet Wars which offers a good degree of skill-depth and is compatible with GVGAI agents.  For future work it would be interesting to compare the skill-depth of this
game with the existing GVGAI games, both for the single and two-player versions.  For the optimisation results
in this paper we used a fixed opponent which means the game can then be treated as a single-player game, the difficulty of
which depends on the strength of the fixed opponent.  Tuning an agent to win a single-player game can be treated
as a standard optimisation problem, though in this case an exceptionally noisy one due to the random initial game states
and the stochastic nature of the agents.  

The results clearly demonstrate how the use of model-based algorithms (both NTBEA and SMAC) were
able to outperform non model-based alternatives, such as a random
mutation hill climber and a simple genetic algorithm.
One of the main messages of this paper is that
games are a natural source of noisy optimisation
problems, and that better performance is often 
obtained by using model-based approaches to
deal with the noise.

NTBEA and SMAC offered similar performance, but NTBEA has the advantage of 
providing more informative output, with detailed statistics for each 
parameter choice in each dimension and in each combination of dimensions
modelled by the n-tuples (see Figure \ref{fig:ntbeaseq}).
NTBEA also offers explicit control over how exploitative versus how explorative
the algorithm should be.

It would be interesting to compare NTBEA to SMAC on optimizing other algorithms or other games, such as tuning an AI agent for General Video Game Playing. Though NTBEA has been applied to tune an MCTS agent for General Video Game AI~\cite{sironi2018self} and General Game Playing~\cite{sironi17online}, it was not compared to SMAC or CMA-ES.  The recently released NeverGrad toolbox
also provides a set of optimisers for further comparisons
\cite{nevergrad}.

The choice of tuples used in NTBEA is naively selected here, and also little effort was made to tune the other parameters of the NTBEA, which are the exploration constant $k$ and the progressive widening parameter epsilon.  The selection of tuples with a-priori knowledge of the agent to be tuned could further increase the efficiency of the optimisation.  Naturally a high-level NTBEA could be used to 
tune an NTBEA running for a specific problem.

% As yet, there is no theoretical analysis on the convergence speed of NTBEA. important point but only if we can flesh it out more

The paper provides further evidence of the
importance of parameter tuning.  The win-rates of the agents evaluated in Figure~\ref{fig:GameScoreIncBuffer}, show how a tuned agent beat an untuned agents 84 games to 16 in one case and 100 games to zero in the other case.  Parameter tuning is important, and
model-based methods often provide the most
efficient approach, especially when the objective function is noisy.

Finally, it is worth emphasizing that the NTBEA does not just
provide useful statistics on the best combinations of parameter,
but actively uses those statistics to inform every decision
about which point in the search space to sample next during a run.

%\end{document}  % This is where a 'short' article might terminate

% \begin{acks}
% This work was funded by the EPSRC Centre for Doctoral Training in Intelligent Games and Game Intelligence (IGGI) EP/L015846/1.
% \end{acks}

\bibliographystyle{aaai}

% \balance
% \bibliographystyle{IEEEtran}

\bibliography{./bibs/jliu,./bibs/gameseminar,./bibs/qmul,./bibs/mariogan,./bibs/planetwars,./bibs/jialinbib}

\end{document}